\documentclass[10pt,twocolumn,letterpaper]{article}

\usepackage{cvpr}              

\usepackage[dvipsnames]{xcolor}

\usepackage[pagebackref,breaklinks,colorlinks]{hyperref}

\usepackage{subfigure}
\usepackage{multirow}

\def\paperID{288} 
\def\confName{3DV\xspace}
\def\confYear{2024\xspace}

\title{Event-Based Visual Odometry on Non-Holonomic Ground Vehicles}

\author{Wanting Xu\thanks{~indicates equal contribution.} \hspace{0.5cm} Si'ao Zhang$^*$ \hspace{0.5cm} Li Cui$^*$ \hspace{0.5cm} Xin Peng \hspace{0.5cm} Laurent Kneip\\
Mobile Perception Lab, ShanghaiTech\\
{\tt\small \{xuwt,zhangsa2023,cuili,pengxin1,lkneip\}@shanghaitech.edu.cn}}

\begin{document}
\maketitle
\begin{abstract}
    Despite the promise of superior performance under challenging conditions, event-based motion estimation remains a hard problem owing to the difficulty of extracting and tracking stable features from event streams. In order to robustify the estimation, it is generally believed that fusion with other sensors is a requirement. In this work, we demonstrate reliable, purely event-based visual odometry on planar ground vehicles by employing the constrained non-holonomic motion model of Ackermann steering platforms. We extend single feature n-linearities for regular frame-based cameras to the case of quasi time-continuous event-tracks, and achieve a polynomial form via variable degree Taylor expansions. Robust averaging over multiple event tracks is simply achieved via histogram voting. As demonstrated on both simulated and real data, our algorithm achieves accurate and robust estimates of the vehicle's instantaneous rotational velocity, and thus results that are comparable to the delta rotations obtained by frame-based sensors under normal conditions. We furthermore significantly outperform the more traditional alternatives in challenging illumination scenarios. The code is available at \url{https://github.com/gowanting/NHEVO}.
\end{abstract}    
\section{Introduction}
Visual Odometry (VO) is a fundamental computer vision problem with important applications in robotics and automotive~\cite{scaramuzza2011visual, nister2004visual}. Research over the past two decades has therefore lead to significant progress and mature visual odometry frameworks for regular cameras~\cite{klein2009parallel, Forster2014ICRA, engel2017direct, murAcceptedTRO2015}. However,
while some of these methods specifically target the application on non-holonomic platforms, current methods face challenges posed by scenarios involving high dynamics and high-dynamic-range conditions. In such situations, regular cameras are easily affected by motion blur or over exposure. To address these challenges and pave the way for further progress in the field of VO, the community has recently started to explore the use of dynamic vision sensors~\cite{weikersdorfer2013simultaneous, censi2014low, mueggler2014event, gallego2018unifying, vidal2018ultimate, zhou2021event, peng2021continuous, 
xu2023tight,
hidalgo2022event}.

\begin{figure}[t]
    \centering
    \includegraphics[width=\columnwidth]{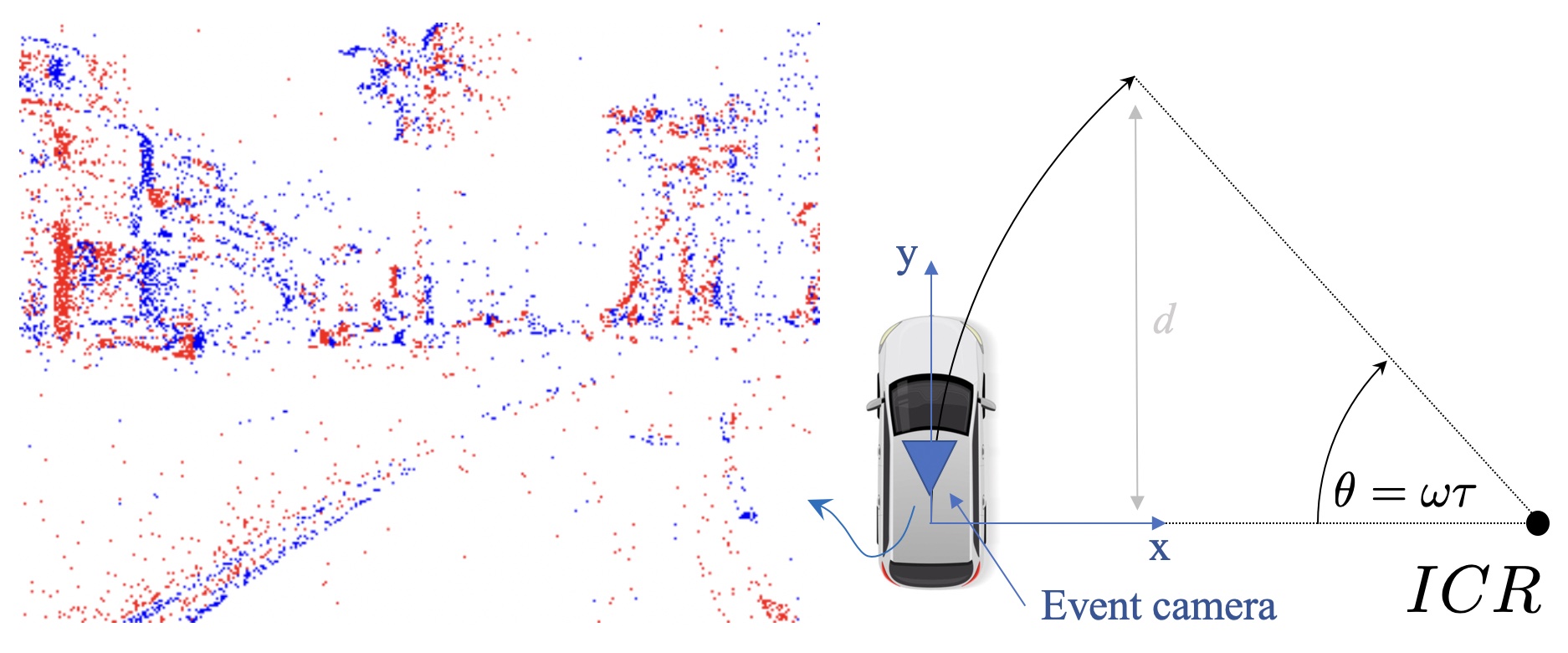}
        \caption{We consider visual odometry with a forward-facing event camera mounted on an Ackermann steering vehicle for which motion can be locally approximated by an arc of a circle about an Instantaneous Centre of Rotation (ICR). We assume constant rotational velocity during the time interval for which events are considered.}
    \label{fig:frontFig}
\end{figure}

Event cameras are innovative visual sensors that offer distinct advantages in high-speed scenarios. Drawing inspiration from the biological retina, each pixel of an event camera functions independently and only responds when a change in brightness level occurs and exceeds a certain threshold. This fundamental circuit-level design provides event cameras with an improved power dissipation, high dynamic range, and reduced response latency, and therefore holds great potential for promoting advancements in visual odometry. Consequently, event-based visual odometry has garnered increasing attention from academia and industry alike~\cite{gallego2020event}. However, the asynchronous mechanism at the pixel level and the relatively immature design of event cameras still pose significant obstacles to the development of event-based visual odometry systems. More specifically, the resulting noise and spatial sparsity in event streams hinder stable and accurate feature extraction as well as the establishment of correspondences over time. The community has therefore converged onto visual-inertial fusion~\cite{zhu17,rebecq2017real,mueggler2018continuous} or even event-frame-inertial fusion strategies~\cite{vidal2018ultimate,hidalgo2022event} for stable motion estimation with event cameras.

The present paper presents a reliable purely event-based visual odometry framework for planar ground vehicles by including a non-holonomic motion model into the estimation. As indicated in Figure~\ref{fig:frontFig}, we use the Ackermann model, a common representation to approximate the instantaneous planar motion of ground vehicles by circular arcs as a function of two degrees of freedom~\cite{scaramuzza2009real,huang19}: the turning radius and the speed of the motion. Combined with the radius, the forward velocity of the vehicle is equivalently expressed by rotational velocity. Based on the assumption of locally constant rotational velocity and turning radius, we introduce a \textit{one-track solver} that is able to initialize the observable part of the instantaneous camera motion from a short, temporal trail of events generated by a single 3D point observed under non-holonomic motion. The algorithm is derived from the standard camera-based one-point algorithm for the two-view scenario~\cite{scaramuzza2009real} and its n-linear extension incorporating both points and lines~\cite{huang19}. As we will show in this work, the extension to event streams alleviates limitations in high-speed and high-dynamic scenarios, while the introduction of the constrained motion model is sufficient to robustify the geometric estimation.
The contributions of our work are summarized as follows:
\begin{itemize}
\item We introduce a continuous-time incidence relation for tracked event corners based on the non-holonomic Ackermann motion model and a locally constant rotational velocity assumption.
\item We employ Taylor expansions to approximate trigonometric functions, resulting in n-linear constraints that depend on higher orders of the rotational velocity. We analyze three distinct expansion levels.
\item We utilize rank minimization to solve the rotational velocity, and eliminate outliers using histogram voting.
\end{itemize}

As demonstrated in our experimental results section, the resulting algorithm leads to accuracy comparable to conventional camera alternatives, and eventually outperforms the latter under challenging illumination conditions. The remainder of this paper is structured as follows. Section~\ref{related_work} offers a literature review. Section~\ref{sec:solver} then provides a brief review of the Ackermann motion model, and presents our proposed n-linear constraint as well as our robust estimation strategy over multiple event tracks.  Section~\ref{sec:experiment} finally presents a thorough evaluation of the algorithm over simulation data, as well as a successful application to challenging real-world cases in which regular camera alternatives fail.

\section{Related Work}
\label{related_work}

The past two decades have seen the development of a large body of visual SLAM solutions. Seminal sparse keypoint based methods have been introduced by Klein et al.~\cite{klein2007parallel}, Mul-Artal et al.~\cite{mur2015orb,mur2017orb}, Campos et al.~\cite{campos2020orb}, and Qin et al.~\cite{qin18vins}. More recently, the community has also presented geometric solutions to the semi-dense~\cite{engel14lsd} or dense case~\cite{min21voldor,koestler21tandem,teed2021droid}. However, these methods all rely on traditional cameras, for which both the establishment of correspondences as well as the geometry are well-understood problems to which many learning and non-learning based solutions exist. Event-based vision is not at a similar level of maturity, and currently still lacks behind even in the development of fundamental geometric pose solvers.

Various solvers for 2D point correspondence-based motion estimation have been proposed based on the epipolar constraint, such as the eight-point solver~\cite{longuet1981computer}, the seven-point solver~\cite{hartley2003multiple}, the six-point solver~\cite{pizarro2003relative}, or---in the minimal case---the five-point solver~\cite{nister2004efficient,stewenius06}. Various specialized solvers for more constrained scenarios have been presented as well, such as solvers for a known directional correspondence~\cite{fraundorfer10}, or a solver for planar motion~\cite{guerrero08planartrifocal}.
In the case of non-holonomic planar motion (i.e. motion adhering to the Ackermann steering model), the problem reduces to only two degrees of freedom. As presented by Scaramuzza et al.~\cite{scaramuzza2009real,scaramuzza2009absolute,scaramuzza2011}, a single correspondence is enough to recover the solution in this case. In~\cite{scaramuzza2009real}, a scale-invariant solution is presented. In~\cite{scaramuzza2009absolute,scaramuzza2011}, a known displacement away from the non-steering axis is used to additionally recover scale. The non-holonomic Ackermann motion model has also been explored for multi-camera systems~\cite{hee2013motion}, or even articulated multi-perspective cameras~\cite{peng2019articulated}. Perhaps most closely related to the present work is the method by Huang et al.~\cite{huang19}, who present a planar tri-focal tensor based~\cite{hartley94,guerrero08planartrifocal} n-linear~\cite{hartley2003multiple} solution able to utilize $n$ measurements of a single line or point captured during a constant velocity arc of a circle in order to accurately determine the non-holonomic motion of the camera. Huang et al.~\cite{huang21bsplines} furthermore introduce a continuous non-holonomic trajectory model for use in back-end optimization. The methods listed here are limited to a traditional constant-framerate sensor operating under mild conditions, hence the present work explores an extension to event cameras.

Most original event-based motion estimation frameworks work in depth constrained scenarios. Weikersdorfer et al.~\cite{weikersdorfer2013simultaneous} propose the first event-based SLAM solution targetting 2D planar motion and planar scenes. Later works often rely on the assumption of a known 3D scene or additional readings provided by a depth camera~\cite{weikersdorfer2014event,censi2014low,mueggler2014event,gallego2016event,bryner2019event,chamorro2020high,zuo2022devo,zuo2024cross}, or are otherwise limited to the pure rotation scenario~\cite{gallego2017accurate,gallego2018unifying}. Zhou et al.~\cite{zhou2021event} present the first stereo event camera visual odometry. An interesting approach that has also found application to event-based motion estimation is given by the contrast maximization method~\cite{gallego2018unifying,gallego2019focus,stoffregen2019event1}. However, the approach is limited to planar homography or pure rotation estimation. Liu et al.~\cite{liu2020globally} and Peng et al.~\cite{peng2021globally} propose globally optimial contrast maximization frameworks. The latter work is particularly interesting, as it also solves ground vehicle motion based on the Ackermann motion model. However, it is using a downward facing camera observing the flat ground plane, and thus still lacks generality.

In order to facilitate motion estimation with event cameras, the community has explored fusion with other sensors. Zhu et al.~\cite{zhu17}, Rebecq et al.~\cite{rebecq2017real}, and Mueggler et al.~\cite{mueggler2018continuous} all explore the combination with interial readings. 
\cite{peng2021continuous, xu2023tight} propose the continuous event-line constraint for camera velocity extraction in line dominated environments, but again use an IMU in order to recover the rotational velocity. Vidal et al.~\cite{vidal2018ultimate} and Hidalgo et al.~\cite{hidalgo2022event} explore the combination with a regular camera. The latter work is the first to propose direct monocular visual odometry using events and frames and photometric bundle adjustment.

Of particular interest to us are visual tracking and mapping frameworks that are able to estimate event camera motion in arbitrary environments and without the assistance of further sensors. Kim et al.~\cite{kim2016real} present a filtering-based approach, while Rebecq et al.~\cite{rebecq2016evo} present a geometric approach alternating between a tracking and a mapping module. Zhu et al.~\cite{zhu2019neuromorphic} propose a framework targeting applications on ground vehicles. However, no specific vehicle motion model is used, and only very few results are being shown.

To the best of our knowledge, our work is the first to include a ground vehicle motion model into single event camera motion tracking and thereby achieve robustness and computational efficiency superior to the existing, more general 6 DoF monocular solutions.
\section{Event-Based Non-Holonomic Solver}
\label{sec:solver}

We now proceed to the derivation of our solver. We start with a review of the Ackermann motion model. Next, we introduce our core contribution, which is a novel algorithm for estimating up-to-scale non-holonomic motion dynamics from a single event trail. More specifically, we propose a novel incidence relation that vanishes for all events generated by the same 3D point under continuous, constant velocity Ackermann motion. To conclude, we adopt rank minimization to solve the constraint and histogram voting over multiple trails to remove outliers.

\subsection{The Ackermann Motion Model}
\label{subsec:ackermann}

As introduced in the original work of Scaramuzza et al.~\cite{scaramuzza2009real, scaramuzza2011}, the motion of a planar Ackermann steering vehicle can be approximated to lie on a circular arc contained in the horizontal plane. The heading of the vehicle furthermore stays tangential to the arc. We adopt the parametrization of Huang et al.~\cite{huang19}, where the body frame is defined such that the $x$-axis points rightward, the $y$-axis forward, and the $z$-axis upward. The $x$-axis is defined to lie at the height of the non-steering back-wheel axis, which is why the Instantaneous Centre of Rotation (ICR) intersects with the $x$-axis. As a result, the instantaneous velocity of the vehicle points along the $y$-axis. A minimal parametrization of a relative displacement is now given by the circle radius $r$ and the inscribed arc-angle $\theta$. The geometry is explained in Figure~\ref{fig:frontFig}. Note that this model is valid under three assumptions: The motion is slip-free, the back-wheel axis is non-steering, and the motion has constant velocity. While the latter assumption does not hold in practice, we demonstrate that the assumption holds sufficiently well over the short time intervals over which the estimation happens.

Let us now define the Euclidean transformation between subsequent frames as a function of our minimal parametrization. As explained in~\cite{huang19}, the relative transformation variables $\mathbf{R}$ and $\mathbf{t}$ that allow us to transform points from the later frame back to the initial frame according to the equation $\mathbf{p}_0 = \mathbf{R}\mathbf{p}_1 + \mathbf{t}$ are given by
%
\begin{equation*}
    \mathbf{R} =
    \begin{bmatrix}
    \cos(\theta) & \sin(\theta) & 0\\
    -\sin(\theta) & \cos(\theta) & 0\\
    0 & 0 & 1`
    \end{bmatrix},\text{ }
    \mathbf{t} = \frac{d}{\sin(\theta)}
    \begin{bmatrix}
    1 - \cos(\theta)\\
    \sin(\theta)\\
    0
    \end{bmatrix},
\end{equation*}
%
\normalsize
where we have replaced $r= \frac{d}{\sin(\theta)}$ to define scale via the forward displacement $d$ along the $y$ axis and avoid numerical issues if $r$ tends to infinity and $\theta$ tends to zero. As can easily be verified by the application of L'H\^opital's rule, we now obtain $\mathbf{t}=[0\text{ }d\text{ }0]^\top$ as $\theta\rightarrow 0$. Note that the convention here is that both $r$ and $\theta$ are positive for a forward right turn (cf. situation illustrated in Figure~\ref{fig:frontFig}), and both $r$ and $\theta$ are negative for a forward left turn.

\subsection{Single Event Trail Constraint}

Let us assume that the constant velocity assumption is satisfied for a short period of time. For each event $\mathbf{e}_i = \{u_i, v_i , t_i, s_i\}$ belonging to a short temporal slice of the space-time volume of events, let $(u_i, v_i)$ define the pixel location of the event on the image plane, $t_i$ its timestamp, and $s_i$ its polarity. Using the continuous-time representation $\theta=\omega \tau$---where $\omega$ represents the rotational velocity of the camera---the relative $\mathbf{R}_i$ and $\mathbf{t}_i$ corresponding to the exact time of event $\mathbf{e}_i$ is given by
%
\begin{equation}
    \mathbf{R}_i =
    \tiny
    \begin{bmatrix}
    \cos(\theta_i) & \sin(\theta_i) & 0\\
    -\sin(\theta_i) & \cos(\theta_i) & 0\\
    0 & 0 & 1
    \end{bmatrix}
    \footnotesize
     =
    \tiny
    \begin{bmatrix}
    \cos(\omega \tau_i) & \sin(\omega \tau_i) & 0\\
    -\sin(\omega \tau_i) & \cos(\omega \tau_i) & 0\\
    0 & 0 & 1
    \end{bmatrix},
    \normalsize
\label{equ:ri}
\end{equation}
%
\begin{equation}
    \mathbf{t}_i 
    \normalsize
    = 
    \tiny
    \frac{d}{\sin(\theta)}
    \begin{bmatrix}
    1 - \cos(\theta_i)\\
    \sin(\theta_i)\\
    0
    \end{bmatrix}
    = 
    \tiny
    \frac{d}{\sin(\omega \tau)}
    \begin{bmatrix}
    1 - \cos(\omega \tau_i)\\
    \sin(\omega \tau_i)\\
    0
    \end{bmatrix}.
    \normalsize
\label{equ:ti}
\end{equation}

Note that $\tau$ here represents a fixed time interval used to again fix the scale as the locally constant turning radius is replaced by $\smash{\frac{d}{sin(\omega\tau)}}$. By again applying L'H\^opital's rule,  we obtain $\mathbf{t}_i = \left[ 0\text{ }\frac{d \tau_i}{\tau}\text{ }0 \right]^\top$ as $\omega\rightarrow 0$, which is as required. Note furthermore that $\mathbf{p}_0 = \mathbf{R}_i\mathbf{p}_i + \mathbf{t}_i$, where $\mathbf{p}_i$ expresses the world point corresponding to $\mathbf{p}_0$ in the displaced frame at time $t_i$. $\tau_i$ represents the time relative to the start time $t_0$ of the considered time interval. That is, for event $\mathbf{e}_i$ with timestamp $t_i$, we have $\tau_i= t_i - t_0$.

\begin{figure}[t]
	\begin{center}
		\includegraphics[width=0.45\linewidth]{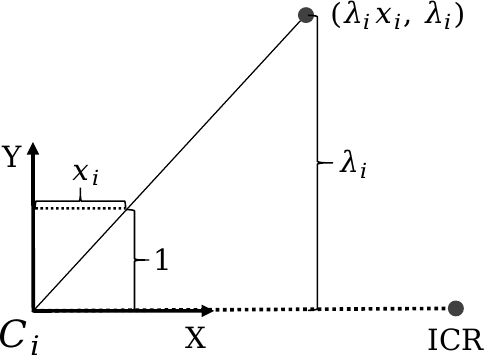}
	\end{center}
	\vspace{-0.1in}
	\caption{The model used in the derivation, where the coordinate system $C_i$ is consistent with the event $\mathbf{e}_i$. Please see text for detailed explanations.}
	\label{fig:model}
\end{figure}

We have $\mathbf{p}_i = \mathbf{R}_i^\top(\mathbf{p}_0 - \mathbf{t}_i)$, and using $\mathbf{p}_i = [p_i^x, p_i^y, p_i^z]^\top$, and by replacing $\mathbf{R}_i$, $\mathbf{t}_i$ with Eqs.~(\ref{equ:ri}), (\ref{equ:ti}), respectively, we obtain

\begin{footnotesize}
\begin{equation*}
\begin{split}
    \begin{bmatrix}
    p_i^x\\
    p_i^y\\
    p_i^z
    \end{bmatrix}
    & = \begin{bmatrix}
    \cos(\omega \tau_i) & -\sin(\omega \tau_i) & 0\\
    \sin(\omega \tau_i) & \cos(\omega \tau_i) & 0\\
    0 & 0 & 1
    \end{bmatrix}
    \\
    & ~~~~~ 
    \Bigg(
    \begin{bmatrix}
    p_0^x\\
    p_0^y\\
    p_0^z
    \end{bmatrix} - 
    \frac{d}{\sin(\omega \tau)}
    \begin{bmatrix}
    1 - \cos(\omega \tau_i)\\
    \sin(\omega \tau_i)\\
    0
    \end{bmatrix}
    \Bigg)
    \\
    & = 
    \begin{bmatrix}
    p_0^x\cos(\omega \tau_i) - p_0^y\sin(\omega \tau_i) - \frac{d}{\sin(\omega \tau)} (\cos(\omega \tau_i) - 1)\\
    p_0^x\sin(\omega \tau_i) + p_0^y\cos(\omega \tau_i) - \frac{d}{\sin(\omega \tau)}\sin(\omega \tau_i)\\
    p_0^z
    \end{bmatrix}.
\end{split}
\end{equation*}
\end{footnotesize}

\normalsize
\par
\setlength{\baselineskip}{12pt}
\noindent Given that this constitutes motion on a ground plane, the $z$ coordinate remains unchanged and we may simply ignore the third equation. Furthermore, for the sake of simplicity, we define the camera frame as being identical with the vehicle frame. Hence, we may directly express $(p_i^x, p_i^y) = (\lambda_ix_i, \lambda_i), i = 1, ..., n$ (illustrated in Fig.~\ref{fig:model}), \ie express the world points in the camera frame as normalized image points multiplied by their depth along the principal axis (in this case the $y$ axis). Note that the entire problem is formulated by projection onto the horizontal plane, and only horizontal bearing measurements are used rather than complete image point measurements (\ie the coordinate along the row dimension is dropped, while $x_i$ is used). The preceding equation can thus be rewritten as \par
\vspace{-0.3cm}
\begin{footnotesize}
\begin{equation*}
\begin{split}
    \lambda_i
    \begin{bmatrix}
    x_i\\
    1
    \end{bmatrix}
    & = 
    \begin{bmatrix}
    p_0^x\cos(\omega \tau_i) - p_0^y\sin(\omega \tau_i) - \frac{d}{\sin(\omega \tau)} (\cos(\omega \tau_i) - 1)\\
    p_0^x\sin(\omega \tau_i) + p_0^y\cos(\omega \tau_i) - \frac{d}{\sin(\omega \tau)}\sin(\omega \tau_i)
    \end{bmatrix}.
\end{split}
\end{equation*}
\end{footnotesize}


The $\lambda_i$ can be effortlessly eliminated by dividing the first row by the second row, resulting in
\small
\begin{equation}
\begin{split}
    x_i
    & = 
    \frac{
    p_0^x\cos(\omega \tau_i) - p_0^y\sin(\omega \tau_i) - \frac{d}{\sin(\omega \tau)} (\cos(\omega \tau_i) - 1)}
    {p_0^x\sin(\omega \tau_i) + p_0^y\cos(\omega \tau_i) - \frac{d}{\sin(\omega \tau)}\sin(\omega \tau_i)}.
\end{split}
\label{equ:xi}
\end{equation}
\normalsize
Following simple derivations, Eq.~(\ref{equ:xi}) can be reformulated into the simple matrix form
%
\begin{equation}
\begin{split}
    \begin{bmatrix}
    a_{i1} & a_{i2} & a_{i3}
    \end{bmatrix}
    \begin{bmatrix}
    p_0^x\\
    p_0^y\\
    d
    \end{bmatrix}
    = 0,
\end{split}
\label{equ:matrixa}
\end{equation}
%
where
\begin{equation}
\begin{split}
    & a_{i1} = -x_i\sin(\omega \tau_i) + \cos(\omega \tau_i),\\
    & a_{i2} = -x_i\cos(\omega \tau_i) - \sin(\omega \tau_i),\\
    & a_{i3} = \frac{x_i\sin(\omega \tau_i) - \cos(\omega \tau_i) + 1}{\sin{(\omega \tau) }}.
\end{split}
\label{equ:ai}
\end{equation}
Equ.~(\ref{equ:matrixa}) represents the constraint between the associated events and motion parameters, and---by stacking multiple measurements from multiple events---it is what we call an \textit{n-linearity}. However, the solution is challenging as the left-hand matrix remains a highly non-linear, trigonometric function of $\omega$.

\subsection{Transformation Into a Polynomial Constraint}

\begin{figure}[t]
    \centering
    \includegraphics[width = 0.45\textwidth]{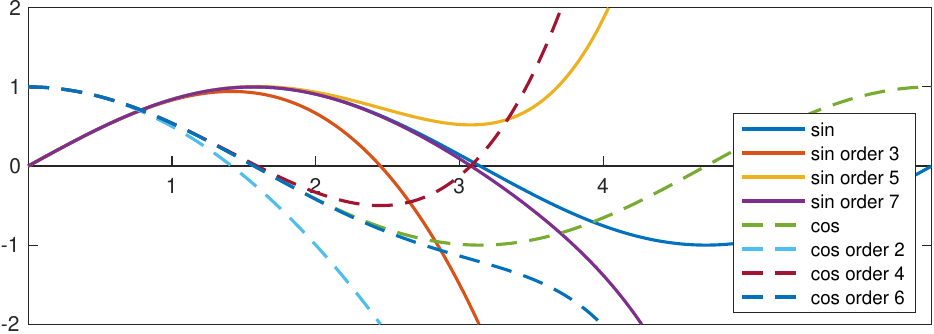}
    \caption{The Taylor series expansion approximations for $\sin$ and $\cos$, the corresponding highest orders are 3, 5, 7 and 2, 4, 6 respectively.} 
    \label{fig:sincos}
\end{figure}

In the following, we adopt a sequence of transformations and approximations in order to transform the n-linearity constraint from Eq.~(\ref{equ:matrixa}) into a simple, uni-variate polynomial. Required by the validity of the constant velocity assumption, $\tau_i$ has to remain small. As a result, $\omega \tau_i$ also remains small. In order to facilitate the subsequent polynomial functions of $\theta = \omega \tau$, we employ Taylor series expansions to approximate trigonometric functions as in
%
%
\begin{equation}
\begin{split}
    &\sin(\theta) \approx \theta - \frac{\theta^3}{3!} + \frac{\theta^5}{5!} + \cdot\cdot\cdot + \frac{(-1)^n\theta^{2n+1}}{(2n+1)!} + \cdot\cdot\cdot,\\
    &\cos(\theta) \approx 1 - \frac{\theta^2}{2!} + \frac{\theta^4}{4!} + \cdot\cdot\cdot + \frac{(-1)^n\theta^{2n}}{(2n)!} + \cdot\cdot\cdot.
\end{split}
\label{equ:sincos}
\end{equation}
The Taylor series approximations for $\sin$ and $\cos$ presented in Fig.~\ref{fig:sincos} correspond to the cut-off orders of 3, 5, 7 and 2, 4, 6, respectively.

Let us denote the \textit{s3c2} algorithm as the one that sets the highest order of the $\sin$ approximation to 3 and the one of the $\cos$ approximation to 2. The \textit{s5c4} and \textit{s7c6} algorithms are defined in a similar way.

Upon substituting Eq.~(\ref{equ:sincos}) into Eq.~(\ref{equ:ai}), with obtain the following form for \textit{s3c2}
%
\begin{equation}
\begin{split}
    & a_{i1} \approx \widetilde{a}_{i1} = x_i(\frac{(\omega \tau_i)^3}{6}-\omega \tau_i) - \frac{(\omega \tau_i)^2}{2} + 1,\\
    & a_{i2} \approx \widetilde{a}_{i2} = x_i(\frac{(\omega \tau_i)^2}{2} - 1) + \frac{(\omega \tau_i)^3}{6} - \omega \tau_i,\\
    & a_{i3} \approx \widetilde{a}_{i3} = \frac{-(\tau_i(- x_i\tau_i^2\omega^2 + 3\tau_i \omega + 6x_i))}{\tau(\tau^2\omega^2 - 6)}.
\end{split}
\end{equation}

For \textit{s5c4} and \textit{s7c6}, due to space constraints, we have placed the results in the supplementary materials. The denominator in $a_{i3}$ can be eliminated by multiplication of both the left and right-hand sides of Eq.~(\ref{equ:matrixa}). We obtain
\begin{equation}
\begin{split}
    \begin{bmatrix}
    b_{i1} & b_{i2} & b_{i3}
    \end{bmatrix}
    \begin{bmatrix}
    p_0^x\\
    p_0^y\\
    d
    \end{bmatrix}
    \approx 0,
\end{split}
\end{equation}
where
\begin{equation}
\begin{split}
    & b_{ij} = c\widetilde{a}_{ij},~j = 1,2,3,
\end{split}
\end{equation}
and for \textit{s3c2}, \textit{s5c4}, and \textit{s7c6}, the value of $c$ is given by
\begin{equation}
  \begin{gathered}
    s3c2:~c = \tau(\tau^2\omega^2 - 6),\\
    s5c4:~c = \tau(\tau^4\omega^4 - 20\tau^2\omega^2 + 120),\\
    s7c6:~c = \tau(\tau^6\omega^6 - 42\tau^4\omega^4 + 840\tau^2\omega^2 - 5040).
  \end{gathered}
\end{equation}

Given $n$ corresponding events, the constraints related to each event can again be assembled into an n-linear problem, resulting in the formulation
\begin{figure*}[t]
    \centering
    \subfigure[]{
    \includegraphics[width = 0.31\textwidth]{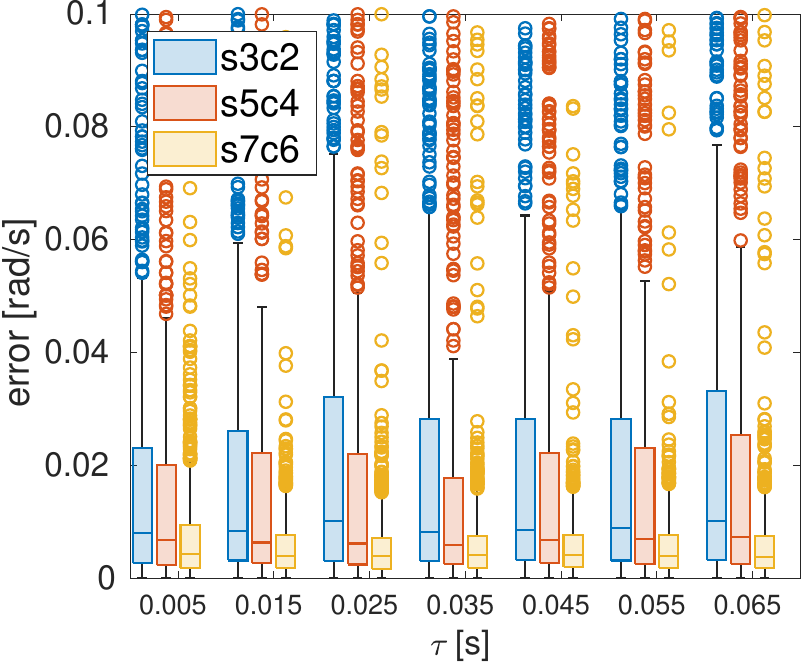}
    \label{fig:tau}
    }
    \subfigure[]{
    \includegraphics[width = 0.31\textwidth]{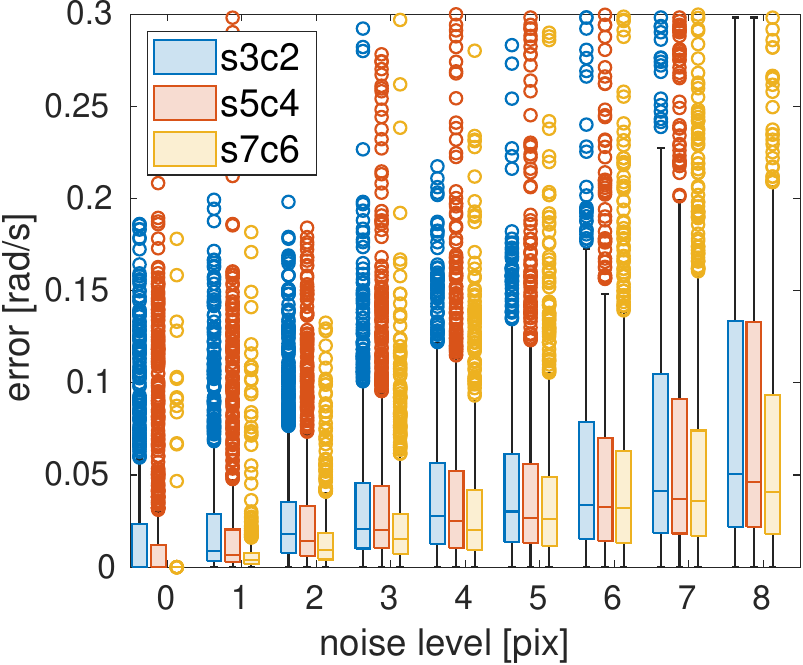}
    \label{fig:noise_level}
    }
    \subfigure[]{
    \includegraphics[width = 0.31\textwidth]{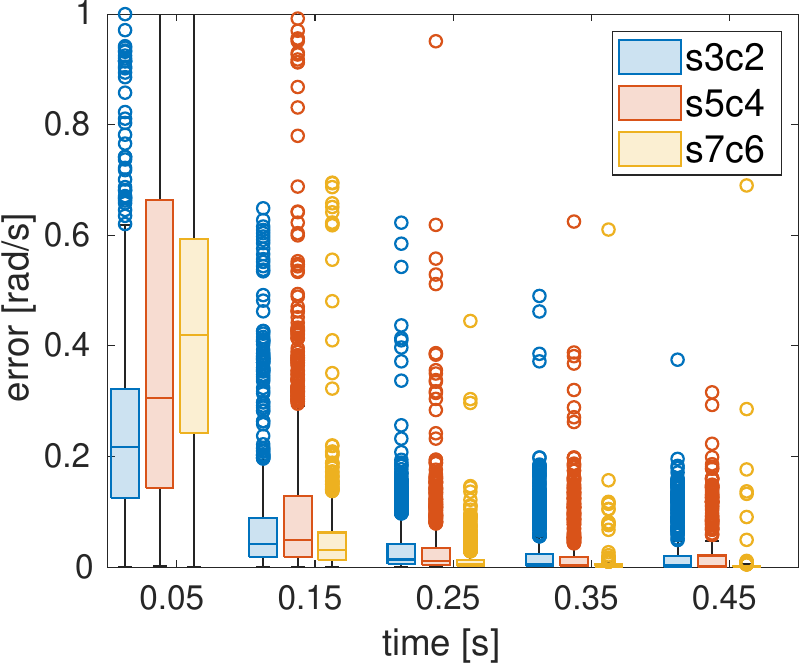}
    \label{fig:time}
    }
    \subfigure[]{
    \includegraphics[width = 0.31\textwidth]{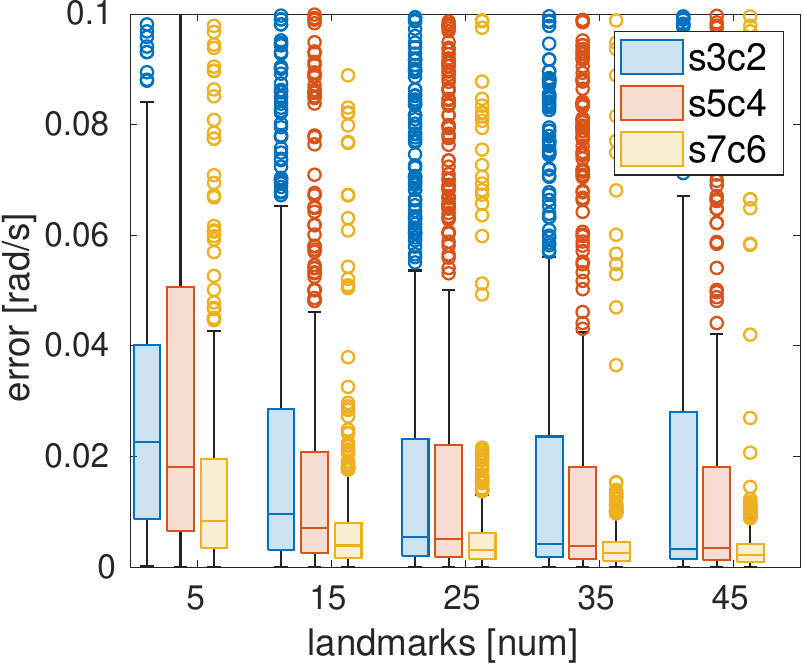}
    \label{fig:landmarks}
    }
    \subfigure[]{
    \includegraphics[width = 0.31\textwidth]{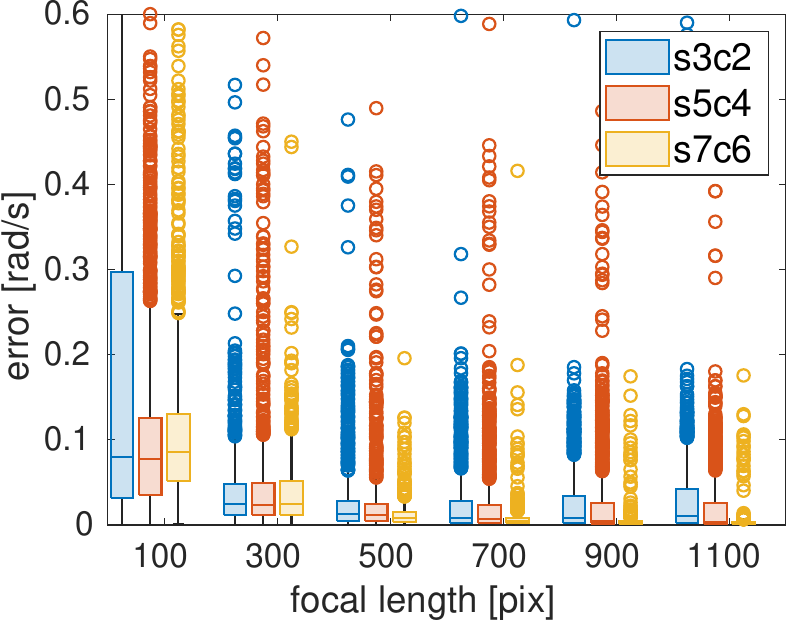}
    \label{fig:focal_length}
    }
    \subfigure[]{
    \includegraphics[width = 0.31\textwidth]{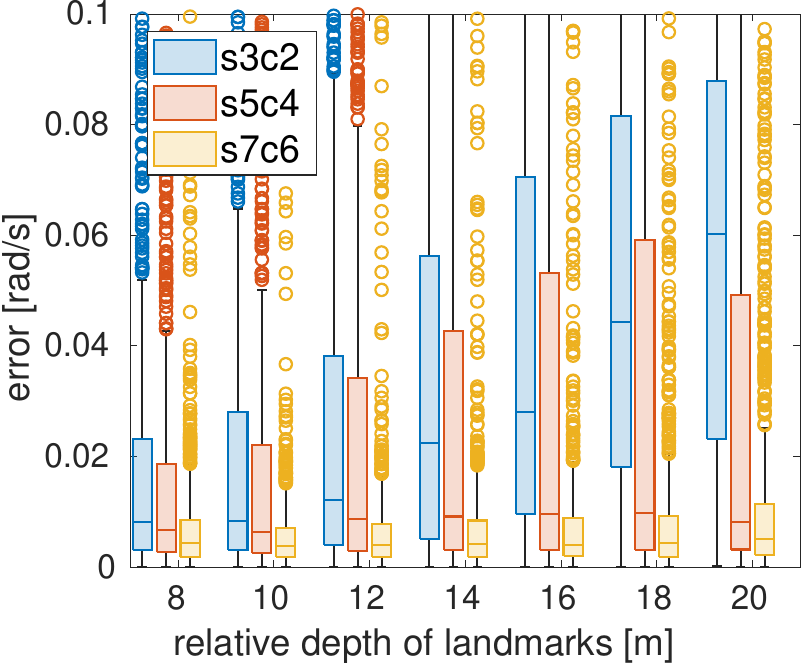}
    \label{fig:relative_depth}
    }
    \vspace{-0.1in}
    \caption{Impact of different factors on the accuracy of the recovered rotational velocity. \textcolor{blue}{s}\textcolor{cyan}{a}\textcolor{blue}{c}\textcolor{cyan}{b} means that the highest order of the Taylor series expansions for \textcolor{blue}{$\sin$} is \textcolor{cyan}{a} and the highest order of the Taylor series expansion for \textcolor{blue}{$\cos$} is \textcolor{cyan}{b}.} 
    \label{fig:simu}
    \vspace{-0.2in}
\end{figure*}

\begin{equation*}
\begin{array}{c}
    \begin{bmatrix}
    b_{01} & b_{02} & b_{03}\\
    \vdots & \vdots & \vdots\\
    b_{i1} & b_{i2} & b_{i3}\\
    \vdots & \vdots & \vdots\\
    b_{(n-1)1} & b_{(n-1)2} & b_{(n-1)3}
    \end{bmatrix}
    \begin{bmatrix}
    p_0^x\\
    p_0^y\\
    d
    \end{bmatrix}\\
\end{array}
\end{equation*}

\begin{equation}
\begin{array}{c}
    = \mathbf{B}[k](\omega)
    \begin{bmatrix}
    p_0^x\\
    p_0^y\\
    d
    \end{bmatrix}
    \approx \mathbf{0},
\end{array}
\label{equ:matrixb}
\end{equation}
where $\mathbf{B}[k](\omega)$ represents a degree-$k$ matrix in $\omega$, and $\mathbf{B}$ refers to a matrix of dimensions ${n\times 3}$.

\subsection{From Rank Minimisation to a Univariate Polynomial Objective}

To obtain a non-trivial solution for $\omega$ in objective~(\ref{equ:matrixb}), it is essential for $\mathbf{B}$ to display rank deficiency. Therefore, the solution for $\omega$ can be facilitated by addressing the rank minimization problem
\begin{equation} \omega_{\mathrm{opt}} = \underset{\omega}{\mathrm{argmin}}~\mathrm{rank}(\mathbf{B}[k](\omega)).
\end{equation}

Given that $\mathrm{rank}(\mathbf{B}) = \mathrm{rank}(\mathbf{B}^\mathrm{T}\mathbf{B})$, the optimisation goal finally becomes
\begin{equation} \omega_{\mathrm{opt}} = \underset{\omega}{\mathrm{argmin}}~\mathrm{rank}(\mathbf{M}[2k](\omega)),
\end{equation}
where $\mathbf{M}[2k](\omega) = (\mathbf{B}[k](\omega))^\mathrm{T}(\mathbf{B}[k](\omega))$ defines a $3 \times 3$ polynomial matrix function of $\omega$. As a positive semi-definite matrix, $\mathbf{M}$'s rank can be minimized by minimising its smallest eigenvalue. The objective becomes
    \begin{equation} \omega_{\mathrm{opt}} = \underset{\omega}{\mathrm{argmin}}~\mathrm{min}(\underset{\lambda}{\mathrm{solve}}(\mathrm{det}(\mathbf{M} - \lambda \mathbf{I}))),
\end{equation}
where $\lambda$ represents the smallest eigenvalue, and $\mathbf{I}$ is a $3 \times 3$ identity matrix. Despite its compact appearance, this objective poses significant optimisation challenges due to its reliance on a repetitive, internal determination of $\mathbf{M}$'s smallest eigenvalue, which is computationally difficult to achieve in closed form. However, it becomes evident that in the perfect noise-free scenario, the rank deficiency condition is met when the smallest eigenvalue of $\mathbf{M}$ at the optimal point simply reduces to zero. Consequently, we approximate $\lambda_{\text{min}} = 0$ and proceed to resolve the final objective
\begin{equation} \omega_{\mathrm{opt}} = \underset{\omega}{\mathrm{argmin}}(\mathrm{det}(\mathbf{M})).
\end{equation}

Observe that this corresponds to identifying the real roots of a univariate polynomial in $\omega$, resolved via Sturm's root bracketing methodology. The determinant polynomials for \textit{s3c2}, \textit{s5c4}, and \textit{s7c6} exhibit an order of 30, 54, and 78, respectively. In practice however, the number of real roots is typically much lower, and much fewer solutions will have to be disambiguated to find the best one.

\section{Experiments}
\label{sec:experiment}
In the following, we will present results for our algorithm collected both in simulation and on real data. In the real-world case, we test on image sequences for both day and night conditions. We employ the histogram voting technique outlined in~\cite{huang19} for handling outlier removal and refining the solution.

\subsection{Experiments on Synthetic Data}
We conduct a variety of experiments with different impact factors, allowing only one parameter to vary within a certain range per experiment. Generally, for the invariant parameters, we configure the noise of each event with a standard deviation of 1 pixel, the time interval length to 0.3 seconds, and the number of observed landmarks to 15. The landmarks have a randomly distributed relative depth between [2, 18] units, with an average value of 10. Additionally, the standard focal length used for the experiments was 700 pixels. In each experiment and setting, we conducted 1000 random experiments. The error $\epsilon$ between the recovered angular velocity $\omega_{rec}$ and the ground truth angular velocity $\omega_{gt}$ is calculated using 
\begin{equation} 
\epsilon = 
| \omega_{rec} - \omega_{gt} |.
\label{error_cal}
\end{equation}

Each experiment evaluates the impact of different highest orders of our Taylor series expansions, including \textit{s3c2}, \textit{s5c4}, and \textit{s7c6}. To recap, in this notation "\textit{s3}" means that the highest order of the Taylor series approximation for $\sin$ is 3 \etc. It is worth noting that attempting to use higher orders of Taylor series expansion beyond \textit{s7c6} would lead to slower calculations despite no further improvements in accuracy. Hence, higher orders are not incorporated into the analysis.

Following is an analysis of each influencing factor: 

\begin{itemize}

\item[$\bullet$] \textit{$\tau$}. \textit{$\tau$} is merely a constant to define the traversed angle within a certain period of time. Together with the sine of that angle and a fixed forward displacement, the constant defines the turning radius. In short, it only impacts on the scale of the problem, and has no implications on the accuracy of the recovered turning rate, which is consistent with the results of Fig.~\ref{fig:tau}.

\item[$\bullet$] \textit{Event noise level}. We conducted performance tests of the algorithm under various noise levels. As shown in Fig.~\ref{fig:noise_level}, we observed that up to noise levels of 5 pixels, the error decreases with an increase in the highest order of the Taylor series expansion. Consequently, \textit{s7c6} achieves the smallest error among the tested options. However, interestingly, at noise levels 6, 7, and 8 pixels, \textit{s5c4} outperforms other configurations. One possible explanation for this phenomenon is that \textit{s7c6}, being of higher order, becomes more sensitive to noise, which adversely affects its performance in high noise conditions. In contrast, \textit{s5c4} strikes a better balance between accuracy and noise sensitivity in those cases.

\item[$\bullet$] \textit{Variation of time interval}. We vary the time interval from 0.05s to 0.45s. As demonstrated in Fig.~\ref{fig:time}, the error decreases as the length of the time intervals goes up. This observation aligns with the analysis presented in \cite{peng2021continuous, xu2023tight}. Nevertheless, it is essential to note that for time intervals exceeding 0.35s, the impact on accuracy is relatively limited. In other words, increasing the time interval beyond this threshold has no significant impact on the algorithm's accuracy.

\item[$\bullet$] \textit{Variation of the number of landmarks}. This parameter indicates the number of landmarks that can be detected from the scene. As seen can be seen in Fig.~\ref{fig:landmarks}, a higher number of landmarks tends to contribute to more accurate results. However, there is a saturation point where a further increase in the number of landmarks has no more impact on accuracy. Furthermore, the superiority of \textit{s7c6} and \textit{s5c4} over \textit{s3c2} implies that higher orders have a positive effect in reducing the error. 

\item[$\bullet$] \textit{Focal length}. As depicted in Fig.~\ref{fig:focal_length}, the error diminishes as the focal length increases. Notably, within the focal length range of 100 to 500, there is a substantial reduction in the error. However, for focal lengths between 500 and 1100, the impact on accuracy becomes less pronounced. This observation holds valuable insights for practical camera applications.

\item[$\bullet$] \textit{Relative depth of landmarks}. For this parameter, the horizontal coordinate represents the mean value of the relative depth distribution, which is uniformly distributed within 8m of that value. In Fig.~\ref{fig:relative_depth}, the errors of \textit{s3c2} and \textit{s5c4} exhibit an increase as the distance increases, and it is particularly pronounced for \textit{s3c2}. However, in the case of \textit{s7c6}, the error remains relatively stable even as the distance increases.

\end{itemize}

Based on the above results, although \textit{s7c6} may exhibit slightly weaker performance than \textit{s5c4} under very high levels of noise, we believe that in practical scenarios such extreme noise levels are not commonly encountered, and that \textit{s7c6} demonstrates better performance in other evaluation metrics. Consequently, for the subsequent real-data experiment, we adopt \textit{s7c6} as a fixed configuration.

\subsection{Experiments on Real-World Data}

To evaluate our algorithm, we conducted tests on KITTI~\cite{geiger2013vision} as well as on our own Self-Collected Data (SCD). The results are compared against the image-based algorithm {\bf 1FPN} by Huang et al.~\cite{huang19}. Note that all qualitative trajectory results are generated by taking additional ground truth scale information into account given that monocular odometry is scale-invariant.

\begin{figure}[t]
    \centering
    \subfigure[Raw Image (KITTI)]
    {
    \includegraphics[width = 0.46\textwidth]{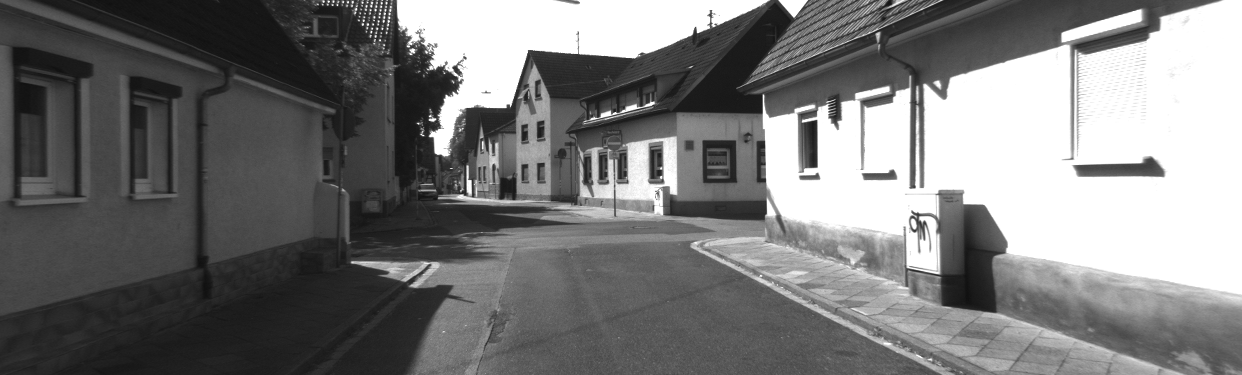}
    \label{fig:raw}
    }
    \subfigure[Cropped Image (KITTI)]{
    \includegraphics[width = 0.22\textwidth]{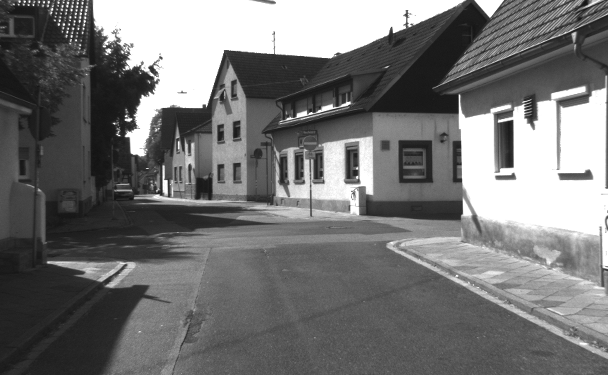}
    \label{fig:crop}
    }
    \subfigure[Dark Night Image (KITTI)]{
    \includegraphics[width = 0.22\textwidth]{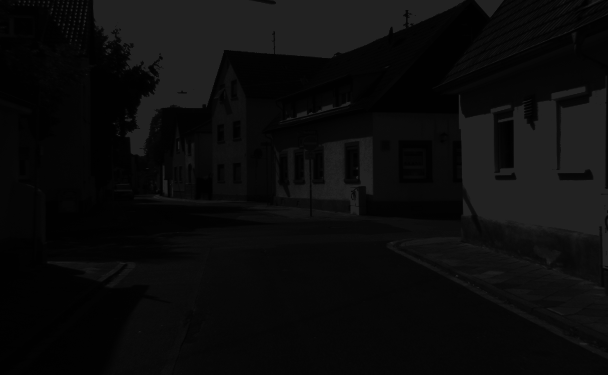}
    \label{fig:dark}
    }
    \subfigure[Events and Corner Tracking (KITTI)]{
    \includegraphics[width = 0.26\textwidth]{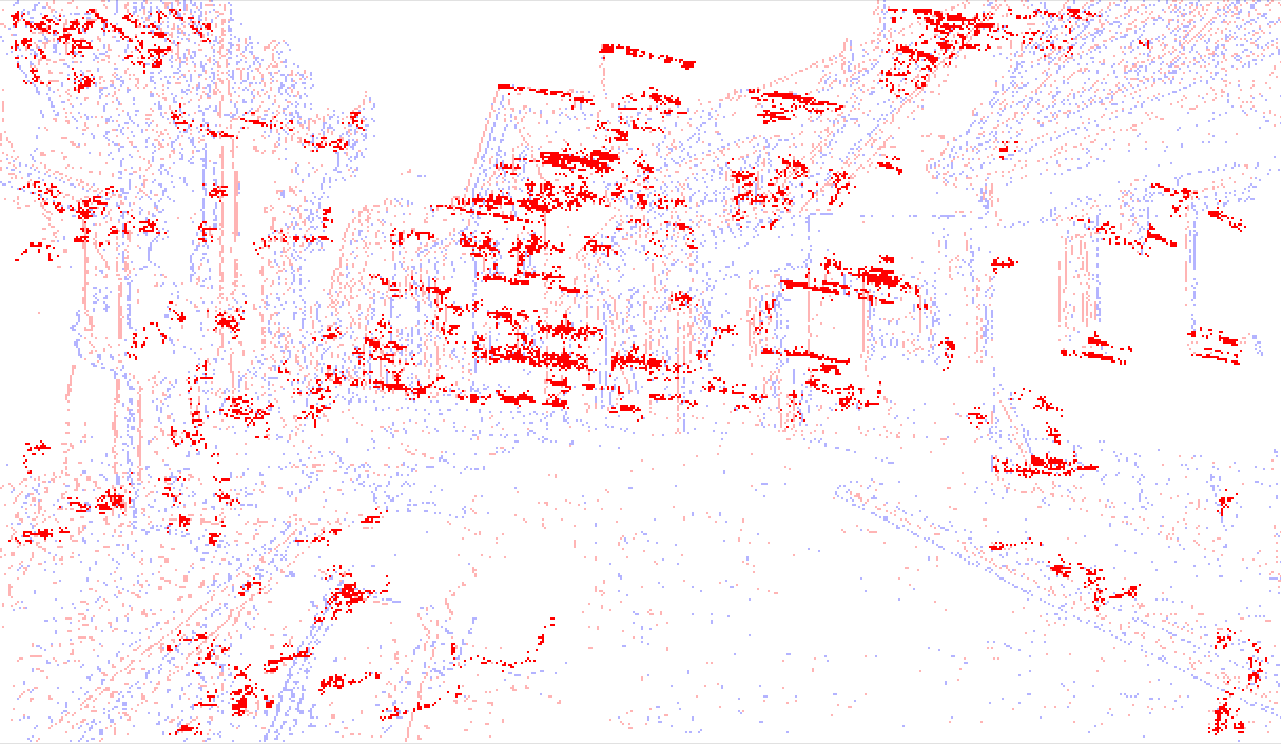}
    \label{fig:event}
    }
    \subfigure[Raw Image (SCD)]{
    \includegraphics[width = 0.18\textwidth]{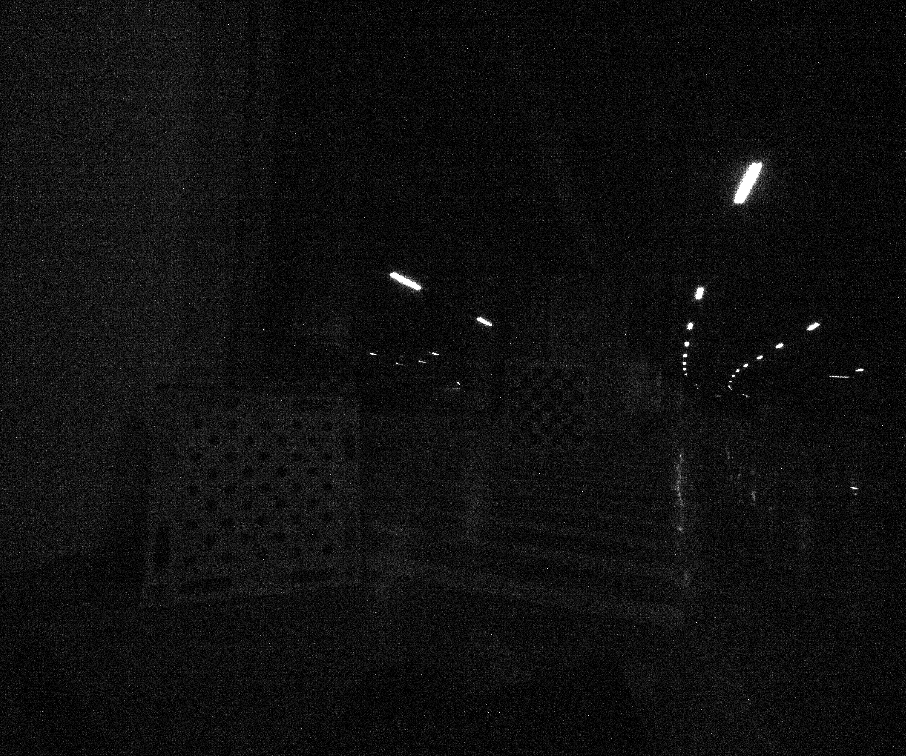}
    \label{fig:real_night}
    }
    \caption{Example images and events with tracked corners used in our real-data experiments. (a), (b), and (c) represent a raw image, a cropped image, and a dark-filtered night image for KITTI, respectively. (d) shows the events generated using vid2e~\cite{gehrig2020video} and demonstrates the corner tracking result. (e) is an example image from our self-collected data at nighttime, on which again almost no features can be extracted.}
    \label{fig:imgs}
    \vspace{-0.2in}
\end{figure}

For KITTI, we test on sequences $0046$, $0095$, and $0104$, respectively. As presently event cameras have relatively low resolution, we cropped the raw KITTI images (Fig.~\ref{fig:raw}) to a resolution of 608 $\times$ 375 (Fig.~\ref{fig:crop}). This step was taken to achieve a better approximation of real-world conditions and enhance computational efficiency. After cropping the image, we retained only the middle part of the raw image, the principal point and focal length are changed accordingly.

KITTI is lacking night images. In order to compare our algorithm with {\bf 1FPN} during both day (D) and night (N) conditions, we first apply a dark filter to the cropped images to approximate nighttime images. The effect of the dark filter can be observed in Fig.~\ref{fig:dark}. It is important to note that---while they may appear differently from real nighttime images---the dark-filtered night images yield similar (or even better) feature extraction results than real night images, hence no disadvantage is given to traditional image-based methods such as {\bf 1FPN}.

The event-based sequences generated on KITTI by application of vid2e~\cite{gehrig2020video} are generated using dark-filtered night images, only. An example result is indicated in Fig.~\ref{fig:event}. 
For event-based feature tracking, we use the work of Alzugaray et al.~\cite{alzugaray2018asynchronous} to detected corner events and return continuous trajectories in the space-time volume.  Trajectories are then evaluated for noise based on node count and duration, resulting in feature trajectories lasting between 0.15 to 0.25 seconds.

SCD was collected at night using a Prophesee Gen3.1 CD event camera, a FLIR Grasshopper3 traditional camera, and a Ouster OS0-128 LIDAR. The LIDAR here is added to obtain ground truth trajectories by running LeGO-LOAM~\cite{legoloam2018}. As shown in Fig.~\ref{fig:real_night}), SCD contains both events and real night time images.

\begin{figure}[t]
	\begin{center}
		\includegraphics[width=1\linewidth]{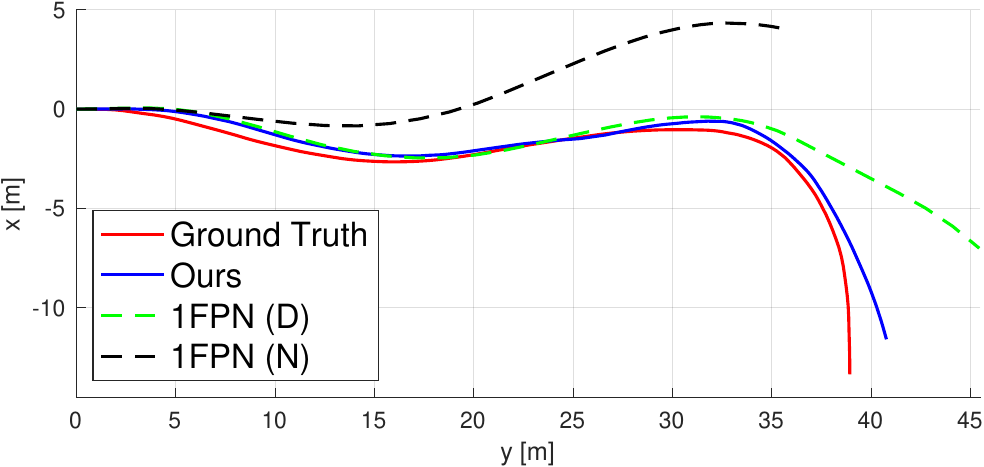}
	\end{center}
	\vspace{-0.2in}
	\caption{Comparison of our event-based method against an image-based method~\cite{huang19} on the KITTI $0046$ sequence, using both the cropped and the dark-filtered night images (D=Day, N=Night).}
	\label{fig:traj}
\end{figure}

\begin{figure}[t]
	\begin{center}
		\includegraphics[width=0.8\linewidth]{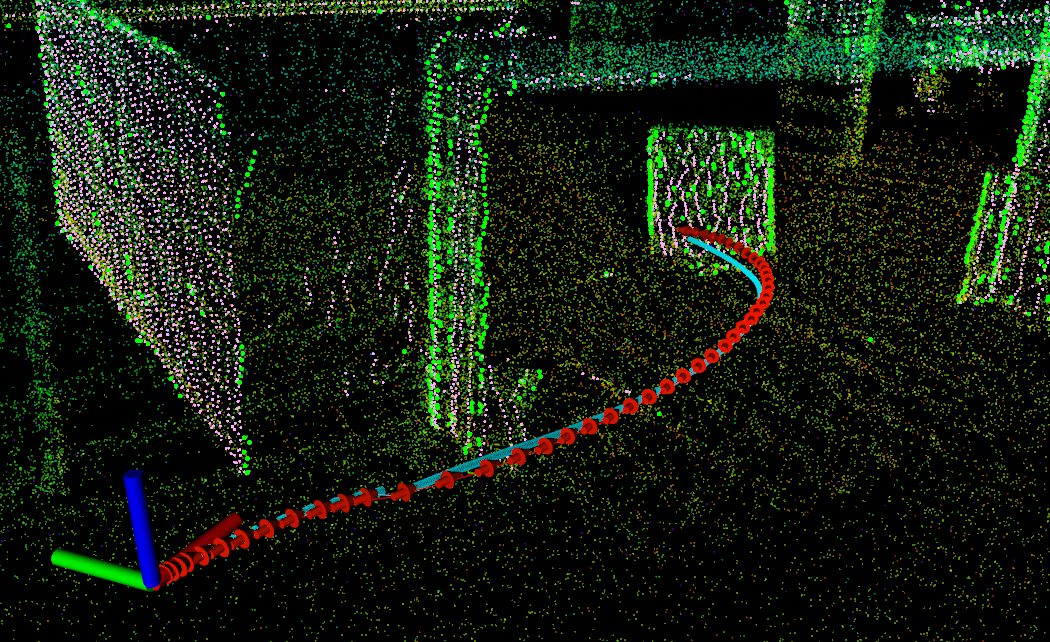}
	\end{center}
    \vspace{-0.4cm}
	\caption{Results of our event-based method (in blue) on SCD with ground truth (in red). The scene was build by LeGO-LOAM. Note that this sequence was collected at night, and the image-based method by Huang et al.~\cite{huang19} does not work.}
 \vspace{-0.4cm}
 \label{fig:traj_selfcollect}
\end{figure}

Fig.~\ref{fig:traj} shows a qualitative trajectory result obtained for KITTI 0046. As can be seen, our results are closest to ground truth and even outperform the traditional image-based alternative applied at day-time ({\bf 1FPN} (D)). The latter algorithm applied at night time ({\bf 1FPN} (N)) exhibits strong drift right from the beginning and fails to complete the trajectory, thus indicating the lack of sufficient accurate features in low-light conditions. Fig.~\ref{fig:traj_selfcollect} displays a result obtained on our night sequences SCD. As can be observed, our method works well while the lack of features in the corresponding night images (cf. Fig.~\ref{fig:real_night}) causes \textbf{1FPN} to fail.

In terms of quantitative comparison, we calculate the errors $\epsilon$ and $\phi$ for the relative rotation angles and translation directions, respectively. The errors are evaluated using both the root mean square (RMS) $\mu$ and median $\nu$. The errors obtained on the different sequences are summarized in Tab.~\ref{tab:errors}, and a runtime efficiency comparison is presented in Tab.~\ref{tab:time}.  As can be observed from Tab.~\ref{tab:errors}, our method almost consistently achieves the lowest error, and significantly outperforms \textbf{1FPN} under dark conditions. As for the time comparison in Tab.~\ref{tab:time}, due to the continuous nature of the event camera, our algorithm reaches higher frequent outputs and achieves real-time performance.

\begin{table}[t]
    \centering
    \caption{Error comparison on different sequences (D=Day, N=Night).
    Note that for SCD, there is only {\bf 1FPN} (N).
    }
    \vspace{-0.05in}
    \begin{tabular}{c|c|cccc}
        \toprule
        \multirow{3}{*}{\bf Seq. } & \multirow{3}{*}{\bf Method} 
        & \multicolumn{4}{c}{{\bf Error}}  \\ 
        \cline{3-6}
        & & $\mu(\epsilon)$ & $\nu(\epsilon)$ & $\mu(\phi)$ & $\nu(\phi)$ \\ 
        \cline{3-6}
        & & [deg] & [deg] & [deg] & [deg]  \\ 
        \midrule
        \multirow{3}{*}{$0046$} &{\bf 1FPN} (D) & 2.7197 & 1.6301 & 2.8090 & 1.2605 \\
        & {\bf 1FPN} (N) & 1.8686 & 1.3022 & 8.1501 & 8.7737 \\
        &{\bf Ours} & {\bf0.9599} & {\bf 0.2154} & {\bf 1.5607} & {\bf 1.2110} \\

        \midrule
        \multirow{3}{*}{$0095$} &{\bf 1FPN} (D) & 0.6467 & 0.5450 & 1.7993 & 1.8080 \\
        & {\bf 1FPN} (N) &- & - & - & -  \\
        &{\bf Ours} &{\bf 0.6187} &  {\bf 0.4024} & {\bf 1.4001} & {\bf 0.9345} \\

        \midrule
        \multirow{3}{*}{$0104$} &{\bf 1FPN} (D) & 0.3732 & 0.2684 & {\bf 0.4679} & 0.4254 \\
        & {\bf 1FPN} (N) &0.5450 & 0.3841 & 0.8870 & 0.7334  \\
        &{\bf Ours} & {\bf 0.0515} & {\bf 0.0357} & 0.5686 & {\bf 0.3608} \\
        
        \midrule
        \multirow{2}{*}{SCD} & {\bf 1FPN} (N) & - & - & - & -  \\
        &{\bf Ours} & {\bf 0.2786} & {\bf 0.1080} & {\bf 2.5317} & {\bf 2.0487} \\
        \bottomrule
    \end{tabular}
    \label{tab:errors}
\end{table}

\begin{table}[t]
    \centering
    \caption{Time comparison on different sequences (D=Day, N=Night). 
    Note that the SCD is collected at night, there is only {\bf 1FPN} (N). "Avg." denotes average calculation time, "Num." is the count of frequent outputs, and "Rem." remarks completion status over the sequence.}
    \vspace{-0.05in}
    \begin{tabular}{c|c|ccc}
        \toprule
        \multirow{3}{*}{\bf Seq. } & \multirow{3}{*}{\bf Method} 
        & \multicolumn{3}{c}{{\bf Time}}  \\ 
        \cline{3-5}
        & & Avg. & Num. & Rem. \\ 
        \cline{3-5}
        & & [ms] & [1] & -\\ 
        \midrule
        \multirow{3}{*}{$0046$} &{\bf 1FPN} (D) & 136.3835 & 41 &  whole\\
        & {\bf 1FPN} (N) &78.0778 & 33 & part \\
        &{\bf Ours} &{\bf 44.7761} & 856 & whole \\

        \midrule
        \multirow{3}{*}{$0095$} &{\bf 1FPN} (D) & 120.4234  & 89 &  whole \\
        & {\bf 1FPN} (N) &- & - & failed \\
        &{\bf Ours} &{\bf 36.0974} & 175 & part \\

        \midrule
        \multirow{3}{*}{$0104$} &{\bf 1FPN} (D) & 118.2199 & 103 & whole \\
        & {\bf 1FPN} (N) &80.2444 & 103 &  whole \\
        &{\bf Ours} & {\bf 16.7670 } & 1381 &  whole\\

        \midrule
        \multirow{2}{*}{SCD} & {\bf 1FPN} (N) &- & - & failed\\
        &{\bf Ours} & 47.1908 & 833 & whole\\
        \bottomrule
    \end{tabular}
    \label{tab:time}
    \vspace{-0.3cm}
\end{table}

\section{Conclusions}
We have introduced a continuous-time adaptation of a frame-based, constant velocity n-view relative displacement solver for Ackermann motion, thereby permitting a robust and accurate solution to monocular event-based visual odometry on a ground vehicle platform. The method clearly outperforms comparable frame-based solutions in difficult, low illumination conditions, and thereby alleviates one of the major disadvantages of traditional vision-based solution for self-driving cars.
\section*{Acknowledgments}
The authors would like to thank the fund support from the National Natural Science Foundation of China (62250610225) and Natural Science Foundation of Shanghai (22dz1201900, 22ZR1441300). We also want to acknowledge the generous support of and continued fruitful exchange with our project collaborators at Midea Robozone.
\clearpage
{
    \small
    \bibliographystyle{ieeenat_fullname}
    \bibliography{main}
}
\clearpage
\setcounter{page}{1}
\maketitlesupplementary

\section{$a_{ij}$ for \textit{s5c4} and \textit{s7c6}}
For \textit{s5c4}, we have
\begin{scriptsize}
\begin{equation*}
\begin{split}
    a_{i1} \approx \widetilde{a}_{i1} = 
    & - x_i(\frac{(\omega\Delta t_i)^5}{120} -
    \frac{(\omega\Delta t_i)^3}{6} + \omega\Delta t_i)\\
    & + \frac{(\omega\Delta t_i)^4}{24} -  \frac{(\omega\Delta t_i)^2}{2} + 1,\\
    a_{i2} \approx \widetilde{a}_{i2} = 
    &  - x_i(\frac{(\omega\Delta t_i)^4}{24} - \frac{(\omega\Delta t_i)^2}{2} + 1)\\
    & - \frac{(\omega\Delta t_i)^5}{120} + \frac{(\omega\Delta t_i)^3}{6} - \omega\Delta t_i ,\\
    a_{i3} \approx \widetilde{a}_{i3} =
    &  \frac{\Delta t_i(x_i\Delta t_i^4\omega^4 - 5\Delta t_i^3\omega^3 - 20x_i\Delta t_i^2\omega^2 + 60\Delta t_i\omega + 120x_i)}{\tau(\tau^4\omega^4 - 20\tau^2\omega^2 + 120)}.
\end{split}
\end{equation*}
\end{scriptsize}
For \textit{s7c6} we obtain
\begin{scriptsize}
\begin{equation*}
\begin{split}
    a_{i1} \approx \widetilde{a}_{i1} = 
    & x_i(\frac{(\omega\Delta t_i)^7}{5040} - \frac{(\omega\Delta t_i)^5}{120} + \frac{(\omega\Delta t_i)^3}{6} - \omega\Delta t_i)  \\ 
    &  - \frac{(\omega\Delta t_i)^6}{720} + \frac{(\omega\Delta t_i)^4}{24} - \frac{(\omega\Delta t_i)^2}{2} + 1,\\
    a_{i2} \approx \widetilde{a}_{i2} = 
    & x_i(\frac{(\omega\Delta t_i)^6}{720} - \frac{(\omega\Delta t_i)^4}{24} + \frac{(\omega\Delta t_i)^2}{2} - 1) \\
    & + \frac{(\omega\Delta t_i)^7}{5040} - \frac{(\omega\Delta t_i)^5}{120} + \frac{(\omega\Delta t_i)^3}{6} - \omega\Delta t_i,\\
    a_{i3} \approx \widetilde{a}_{i3} = 
    & \frac{-(\Delta t_i(- x_i\Delta t_i^6\omega^6 + 7\Delta t_i^5\omega^5 + 42x_i\Delta t_i^4\omega^4 - 210\Delta t_i^3\omega^3}{\tau(\tau^6\omega^6 - 42\tau^4\omega^4 + 840\tau^2\omega^2 - 5040)}\\
    & \frac{-(\Delta t_i(- 840x_i\Delta t_i^2\omega^2 + 2520\Delta t_i\omega + 5040x_i))}{\tau(\tau^6\omega^6 - 42\tau^4\omega^4 + 840\tau^2\omega^2 - 5040)}.
\end{split}
\end{equation*}
\end{scriptsize}

\end{document}